\documentclass[conference,final,letterpaper]{ieeeconf}

\IEEEoverridecommandlockouts

\usepackage{microtype}

\usepackage{chngcntr}
\counterwithout{paragraph}{subsubsection} 
\counterwithin{paragraph}{subsection} 

\usepackage{cite}
\usepackage{amssymb}
\newcommand{\argmin}{\operatornamewithlimits{argmin}}
\newcommand{\argmax}{\operatornamewithlimits{argmax}}
\usepackage{color}

  \usepackage[pdftex]{graphicx}
  \DeclareGraphicsExtensions{.pdf,.jpeg,.png}

\usepackage[cmex10]{amsmath}

\begin{document}
\title{\LARGE \bf
Computational Models of Tutor Feedback in Language Acquisition}

\author{Jens Nevens$^{1}$ and Michael Spranger$^{2}$%
\thanks{$^{1}$Vrije Universiteit Brussel, Brussels, Belgium}%
\thanks{$^{2}$Sony Computer Science Laboratories Inc., Tokyo, Japan}
}


\maketitle
\thispagestyle{empty}
\pagestyle{empty}

\begin{abstract}
This paper investigates the role of tutor feedback in language learning using computational models. We compare two dominant paradigms in language learning: \emph{interactive learning} and \emph{cross-situational learning} - which differ primarily in the role of social feedback such as gaze or pointing. We analyze the relationship between these two paradigms and propose a new \emph{mixed} paradigm that combines the two paradigms and allows to test algorithms in experiments that combine no feedback and social feedback. To deal with mixed feedback experiments, we develop new algorithms and show how they perform with respect to traditional \emph{knn} and prototype approaches. 
\end{abstract}

%
\IEEEpeerreviewmaketitle

\section{Introduction}
Understanding lexicon learning in children, robots and artificial agents is one of the prime targets for developmental robotics as is evidenced by a steady stream of models, experiments and publications \cite{roy2002learning,fontanari2009cross,kachergis2012cross}. A particular focus has been on agent-based models that study language learning using artificial software agents where one agent (the tutor) is teaching another agent (the learner) a particular language. One of the key factors that influences the complexity of the learning task is the presence or absence of social feedback. Consequently, work on lexicon learning traditionally falls into two sets of models: \emph{cross-situational} learning models and \emph{interactive learning}.

In \emph{cross-situational} learning models, the learner sees a context of objects, observes a word (or multiple) and over many interactions has to figure out which object the word refers to. The key problem solved by the learner in these models is which word refers to which object (or which aspects of objects). This paradigm has been studied with infants \cite{smith2008infants}, adults \cite{yu2007rapid,chen2015cross} and in simulation \cite{siskind1996computational, smith2006cross, fazly2010probabilistic, frank2007bayesian}. \emph{Hypothesis} based models store a single hypothesized referent, until evidence for the contrary is presented \cite{medina2011words,trueswell2013propose}. However, \emph{associative} models (e.g. \cite{fazly2010probabilistic}), using statistical analysis of words and objects, align better with empirical findings \cite{kachergis2012cross,cassani2016constraining}. 

The second type of models is called \emph{interactive learning}. Here the learner not only observes a number of objects and an utterance but additionally receives social feedback (often in the form of pointing or gaze) that clearly restricts the possible referents of the word to one object in the context. Interactive learning models have been proposed for the learning of spatial language \cite{spranger2013acquisition}, color \cite{bleys2009grounded} and other domains. They are also a prime paradigm for models of lexicon evolution \cite{steels2001language,bleys2015language,spranger2016evolution}.

Almost all models in lexicon learning fall into either of the two categories. However, very little work has been done to compare these two paradigms systematically, except for \cite{belpaeme2012word} who show that cross-situational learning is slower than interactive learning, because the social feedback in interactive learning helps the learner. \cite{belpaeme2012word} compare two algorithms: \emph{K Nearest Neighbor} (KNN) for cross-situational learning and \emph{prototype estimation} for interactive learning. Importantly, real life interactions between caregiver and child probably do not fall into either of the two categories but are really combinations of social feedback, little or no feedback and/or unreliable feedback. Consequently, learning algorithms must be able to deal with situations where there is feedback interleaved with interactions where there is no feedback. 

In this paper, we extend traditional approaches such as KNN and prototype learning to mixed interaction scenarios that allow to control the amount of social feedback in each interaction and we propose a number of algorithms for solving mixed cases including the extreme cases of cross-situational learning and interactive learning. We quantify how our algorithms perform with respect to learning success.
 
The paper proceeds as follows. We first introduce the experimental paradigm and discuss its relation to \emph{interactive} and \emph{cross-situational} learning. We then introduce four algorithms and test whether and how they deal with cross-situational, interactive and mixed learning scenarios. We quantify the performance of the algorithms in experiments and draw some conclusions about the impact of tutor feedback in language learning.

\section{Experimental Setup}
\label{sec:setup}
In this paper, we examine the role of tutor feedback in agent-based models of continuous-valued meaning lexicon learning in consecutive interactions between a learner and a tutor agent. The tutor is given a lexicon of word meaning mappings before the experiment, and the learner has to pick up these words and their meaning in situated interactions. Each experiment has two phases: \emph{training} and \emph{testing}. Both take place in the same environment.

\subsection{World}
A world consists of a pool of objects $O$ of size $n$ (also called the world size). Each object is made up of features. In the experiments reported here, the objects consist of three continuous features $O \subset \mathbb{R}^d$ with $d=3$. $\mathbb{R}^3$ can represent various sensorimotor spaces, e.g. three color channels of the YUV color space. The world is initialized at the start of each experiment by uniformly sampling objects from $\mathbb{R}^3$. Every interaction between tutor and learner takes place in a different environment -- the \textit{context} $C$. The context consists of a set of $m$ objects, randomly drawn from $O$ ($C \subset O$, $m<n$).

\subsection{Training} 
In training, the tutor teaches the learner the language by providing examples of words in particular environments. Training interactions happen in the following way:

\begin{enumerate}
\item Choose a set of objects $C \subset O$ of size $|C| = m$ ($m<n$)
\item The tutor chooses a random object from the context, called the \textit{speaker topic} $o_s\in C$.
\item The tutor chooses a word $w \in W$ to describe or distinguish the topic. How this is done depends on the tutor's production strategy (explained later). The tutor passes the word to the learner.
\item Depending on the particular paradigm the learner might or might not receive pointing to the topic from the tutor. The frequency of pointing is determined by a binomial distribution $B(1,f)$ with $f\in [0,1]$. $f$ is a parameter of the experiment.
\item Word, context and pointing feedback (if available) are used by the learner to update his internal representations.
\end{enumerate}

\subsection{Testing} The goal of the training phase is for the learner to become successful in interacting with the tutor. We check the performance of the learner using \emph{testing} interactions:

\begin{enumerate}
\item Randomly assign the roles of speaker and hearer to the learner and the tutor.
\item Choose a set of objects in the context $C \subset O$.
\item The speaker chooses a random object from the context, called the \textit{speaker topic} $o_s\in C$.
\item The speaker chooses a word $w \in W$ to describe or distinguish the topic and passes it to the hearer.
\item The hearer interprets the word $w$ and identifies the topic $o_h \in C$ of the utterance. The hearer points to $o_h$.
\item The interaction is a success iff $o_s=o_h$
\end{enumerate}

In testing, the learner can be both speaker and hearer. The learner has to show how well he is capable of understanding the tutor (interpretation) and how well he is understood by the tutor (production). Success is confirmed through pointing. That is, a test interaction succeeds if both speaker and hearer think that the word chosen by the speaker refers to the same object.

\subsection{Tutor language} 
An important aspect of these experiments is how the tutor represents the target language that needs to be acquired by the learner. We here take a prototype approach. The tutor randomly generates $|W|=t$ words. The meaning of each word is a prototype $p_w\in P \subset \mathbb{R}^3$. The prototypes are randomly generated from a uniform distribution over $\mathbb{R}^3$.

In a particular context $C$ and some randomly chosen topic $o_s \in C \subset \mathbb{R}^3$, the tutor as the speaker chooses the prototype closest to the topic and most far away from the closest other object in the context as in the following formula:
\begin{eqnarray}
\label{eqn:tutor-prod}
p &=& \argmax_{p \in P}[ \min_{o \in C \setminus \{o_s\}} \operatorname{d}(p,o) - \operatorname{d}(p,o_s) ]
\end{eqnarray}
with $P$ all prototypes known by the tutor, and $\operatorname{d}$ being the Euclidian distance ($d(x,y)\geq 0$). Each prototype corresponds to exactly one word (bijective mapping). So choosing a prototype immediately results in the tutor knowing which word to utter to the learner. This strategy is called \emph{discriminative production}, because the tutor chooses words that are most discriminative -- optimizing distance between the topic and other objects in the context.

The tutor can also be hearer in test interactions. In interactions where the learner is producing words as the speaker, the tutor hears a word $w$ and retrieves the prototype $p_w$. He then has to decide which of the objects in the context the word refers to. He does this by retrieving the object in the context $o\in C$ with the shortest distance to $p_w$:
\begin{eqnarray}
\label{eqn:tutor-int}
o_h &=& \argmin_{o \in C} \operatorname{d}(p_w,o)
\end{eqnarray}
with $\operatorname{d}$ being the Euclidian distance. 

\subsection{Analysis of the setup}
Often work on language learning takes either an interactive (always feedback) or a cross-situational learning (no feedback) point of view. Our setup allows to test algorithms with respect to various forms of feedback. If the tutor always points to the object he has in mind then we are in an interactive learning paradigm ($f=1$). If the tutor does not point to the topic ($f=0$) then we are in a cross-situational learning paradigm. If the tutor sometimes points and sometimes does not point ($0 < f < 1$) then we are in a \emph{mixed} learning environment.

We evaluate various learning algorithms in the same way using the test interactions. However, for successful test interactions, the representations of tutor and learner do not have to be the same. In fact, our formulation of the learning problem turns word learning into a multi-class, online machine learning problem with noisy labels (depending on pointing frequency). This means that we can use and test a variety of machine learning algorithms.

Notice also that we are in a continuous meaning domain with $O \in \mathbb{R}^3$ especially if the tutors prototypes $P \neq O$. The tutor lexicon can be larger or smaller than the actual number of objects that the learner encounters. The tutor lexicon does not aim to uniquely identify each object, but instead represents more generic colour categories. Notice also that depending on the training time and the tutor lexicon size, the learner may not have encountered all objects and/or words in training.

The experiments have a number of parameters: $f \in [0,1]$ - pointing frequency, $d=3$ - dimensionality of the object space, $n$ - the number of objects in the world, $m$ - the context size for training and testing interactions, $W$ and $P$ - the words and prototypes (number and exact location in $\mathbb{R}^d$) that have to be learned.

\section{Learning algorithms}
The goal of the learner is to become a proficient speaker and listener with respect to the target language of the tutor. We developed four algorithms able to deal with mixed feedback scenarios, each of them using a different strategy to estimate the tutor language. In particular, we employ a variant of $k$ Nearest Neighbor (KNN) adjusted to deal with mixed cases. We also propose three new prototype-based (centroid) algorithms. Both KNN  and centroid algorithms have been used for language learning in continuous semantic domains. KNN is the only tested algorithm for continuous meaning cross-situational learning \cite{belpaeme2012word}. Centroid algorithms are the goto method for interactive learning \cite{bleys2009grounded,spranger2013acquisition,spranger2016referential}. 

\subsection{k Nearest Neighbor (KNN)}
One algorithm that has been applied to continuous meaning domains \cite{belpaeme2012word} is $k$-Nearest Neighbor (KNN) classification technique. This is a non-parametric technique that is one of the simplest and most effective methods available in Machine Learning. The learner stores all samples of naming events without computing any abstractions.

\paragraph{Representation} The learner keeps a word-object memory that contains the observed objects $o_i \in \mathbb{R}^3$ and words $w_i \in W$ associated with these objects.

\paragraph{Learning} In training, the learner observes a word $w$, the context $C$ and possibly pointing. For KNN the learner first computes the \emph{topic set} $T$
\begin{eqnarray}
\label{eq:topic-set}
T&=&\begin{cases}C & \text{if no pointing} \\ \{o_s\} & \text{if pointing and $o_s$ topic pointed to}\end{cases}
\end{eqnarray}
and then simply adds a sample for each $o_i \in T$ to his word-object memory. If $|T| = 1$ (there was pointing), then we add $o_s$ $m$ times to the word-object memory with $m = |C|$. 

\paragraph{Production and Interpretation} In testing, the learner has to produce and interpret words. In order to produce a word for the topic $o_s$, the learner finds the $k$ nearest neighbors to the topic using the samples in his word-object lexicon. From these $k$ word-object entries, the learner chooses the word that occurs most frequently. Interpretation works similarly. To interpret a given word $w$, the learner will take the $k$ nearest neighbors in its word-object memory for each object in the context and classify each object as the most occurring word in its $k$ neighbors. The interpreted topic is the object which has been classified with $w$. If multiple objects are classified with $w$, one of the objects is chosen at random. 

\paragraph{Parameters} This method has the hyper-parameter $k$. Here we use $k=30$.

\subsection{Prototype Estimation (PE)}

In interactive learning researchers often propose to use prototypes for the learner's lexicon\cite{bleys2009grounded,spranger2013acquisition,spranger2016referential} and to update these prototypes iteratively as new samples of word object mappings are collected by the learner.

\paragraph{Representation} The learner stores prototypes $p\in \mathbb{R}^3$. 

\paragraph{Learning} The prototype $p_w$ is updated when observing the word $w$ in a training interaction using the previous prototype $p^t_w$ and the topic set $T$ (see Equation \ref{eq:topic-set}) with the following equation.
\begin{eqnarray}
\label{eq:lg-update}
p_w^{t+1} &=& (1 - \alpha)p_w^{t} + \alpha \frac{1}{|T|}\sum_{o \in T} o
\end{eqnarray} 
This update shifts the prototype of the learner $p$ more towards the encountered objects $T$ mediated by $\alpha \in [0,1]$. If the learner has not heard $w$ before, then we assume $p^t_w=0 \in \mathbb{R}^3$ and $\alpha = 1$.

\paragraph{Parameters} This method has the hyper-parameter $\alpha \in [0,1]$ with typical $\alpha=.05$.

\paragraph{Production and Interpretation} Since the learner has the same representation as the tutor, we can apply the same production and interpretation equations as the tutor (Equations \ref{eqn:tutor-prod} and \ref{eqn:tutor-int}).

\paragraph{Behavior} In interactive learning ($f=1,|T|=1$), the algorithm behaves like a vanilla centroid model with weighted update. The equation then reduces to 
\begin{eqnarray}
\label{eq:lg-update}
p_w^{t+1} = (1 - \alpha)p_w^{t} + \alpha o_s.
\end{eqnarray} 
which is an  update rule proposed in \cite{bleys2009grounded}. The update will incorporate all objects in the context in interactions without feedback. The intuition behind this learning rule is the following. Imagine a series of interactions with a context of two objects, one always being the same topic, the other being a randomly chosen distractor object. Suppose further the tutor is always using the same word $w$. By always adding objects to $p_w$ of which one is the correct object, and averaging over them, the prototype for this word should converge towards the prototype known by the tutor, because the distraction objects cancel each other out -- if they are generated uniformly.

\subsection{Averaging Prototypes (AP)}
This algorithm is a variant of PE that keeps around full sets of samples for words and estimates prototypes based on this set (not just the last sample as in PE).

\paragraph{Representation} The learner stores prototypes $p\in \mathbb{R}^3$. He also keeps a list of samples $S$ of all past context indexable by the word $w$ as $S_w$. 

\paragraph{Learning}  Suppose the learner has observed the context $C$, he can then compute the topic set $T$ (as in Equation \ref{eq:topic-set}). The learner estimates the new prototype $p_w^{t+1}$ using $T$ and $S_w$ which is all samples for word $w$. 
\begin{eqnarray}
\label{eq:csl-avg-update}
S^{t+1}_w &= S^{t}_w \odot T\\
p^{t+1} &= \frac{1}{|S^{t+1}_w|}\sum S^{t+1}_w
\end{eqnarray} where $\odot$ denotes the concatenation of new samples $o_1,...,o_m$ to the list of samples $S_w$. If the learner has not heard $w$ before, then we assume $S_w^t$ is empty.

\paragraph{Production and Interpretation} Since the learner has the same representation as the tutor, we can apply the production and interpretation of Equations \ref{eqn:tutor-prod} and \ref{eqn:tutor-int}.

\paragraph{Parameters} This method has no hyper-parameters.

\paragraph{Behavior} AP behaves similar to PE but with the difference being that all samples have equal influence, whereas in PE samples are weighted by recency. 

\subsection{Co-occurrence Weighted Prototypes (CWP)}
CWP is another algorithm that uses prototypes as basic representation. However, here we add additional information from co-occurrence of words and objects into the update rule. 

\paragraph{Representation} The learner estimates prototypes $p\in \mathbb{R}^3$ (centroids) from a set of samples by accumulating examples from the tutor. He also keeps a matrix $CC$ for tracking co-occurrences between words $w$ and object $o$ with $CC \in W\times O$. $CC$ is initialized all zeros. 

\paragraph{Learning} Suppose the learner is in context $C$ and has observed $w$. He then estimates topic set $T$ as in Equation \ref{eq:topic-set}. The learner then updates $CC$ and $p_w$ using the following two Equations
\begin{eqnarray}
\label{eq:csl-weighted-update-co}
CC_{w,o_i}^{t+1}&=&\begin{cases}
CC_{w,o_i}^t + 1 & ~\text{iff}~o_i\in T\\
CC_{w,o_i}^t & ~otherwise
\end{cases}\\
\label{eq:csl-weighted-update-p}
p_w^{t+1}&=&(1-\alpha)p_w^{t} + \alpha \sum_{o_i \in T}{\beta_i o_i}\\
~\text{where}~\beta_i &=& \frac{CC^{t+1}_{w,o_i}}{\sum_{o \in T}CC^{t+1}_{w,o}}
\end{eqnarray}
with $\beta$ being a normalization factor, $\alpha\in [0,1]$. We set $\alpha = 1.0$ if the learner hears word $w$ for the first time. The rule weights the impact of each object in the topic set on the prototype update by how often word/prototype and object co-occurred. Objects that co-occurred more frequently with the word $w$ and are in the topic set $T$ have more influence on the prototype $p_w$. $\alpha$ controls how much influence each interaction has on the prototype update.

\paragraph{Production and Interpretation} Since the learner has the same representation as the tutor, we can apply the same production and interpretation (Equations \ref{eqn:tutor-prod} and \ref{eqn:tutor-int}).

\paragraph{Behavior}
In interactive learning ($f=1.0$), $T=\{o_S\}$ and $|T| = m = 1$. It follows that $\beta = 1$ which means that this rule in interactive learning is the same as for PE. In cross-situational learning, $f=0.0$, $m\neq 1$ and consequently the impact of each object in the context is weighted by its overall co-occurrence with word $w$ with respect to all other objects in the context. Suppose that we are in a context $C=\{o_1,o_2\}$, and furthermore suppose that $o_1$ has co-occurred always with $w$ and $o_2$ never, then $p_w$ will be updated more with $o_1$. However, if $o_1$ and $o_2$ have co-occurred with $w$ roughly the same time, then the update of $p_w$ is averaged over $o_1$ and $o_2$.

\section{Results}
\subsection{Interactive Learning ($f=1$)}
\label{sec:interactive-learning}

\begin{figure}[!t]
	\centering
	\includegraphics[width=2.4in]{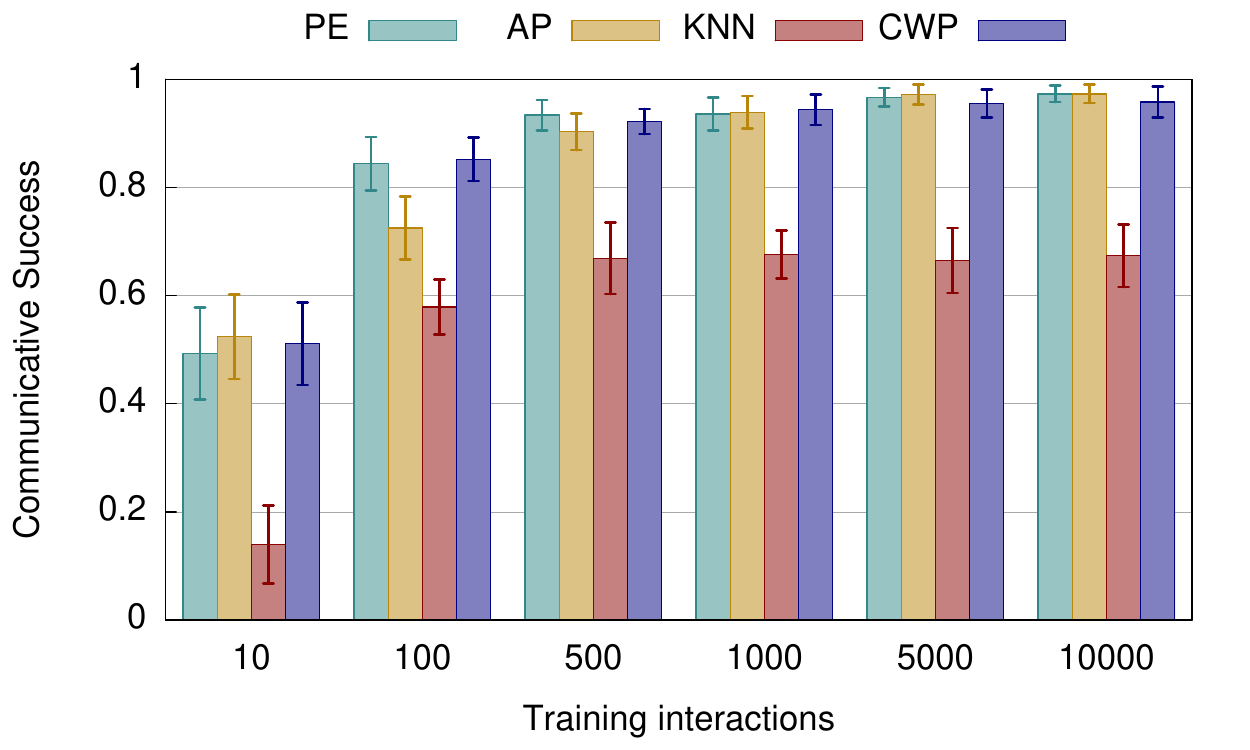}
	\caption{Communicative success for increasing training interactions using the interactive learning paradigm ($f=1$). World size = 32, context size = 4, tutor lexicon size = 50, 100 testing interactions, 20 repetitions. Error bars show standard deviation.}
	\label{fig:IL-dynamics}
\end{figure}

Figure~\ref{fig:IL-dynamics} shows the communicative success obtained in testing interactions for the various learning algorithms, all using the interactive learning paradigm ($f=1$). The prototype-based algorithms (PE, AP and CWP) achieve high levels of communicative success. Also, the communicative success increases very quickly (50\% after only 10 interactions). This points to the fact the learner acquires the lexicon very rapidly. We argue that this rapid learning is due to the availability of social feedback which allows these algorithms to update the prototypes using a single object.

This high level of success is not achieved by the KNN algorithm. One reason for this might be the difference in lexicon representation between tutor and learner when using this algorithm. However, as we will discuss later on, the production strategy of the tutor also plays a role.

\subsection{Cross-Situational Learning ($f=0.0$)}
\label{sec:cs-learning}

\begin{figure}[!t]
	\centering
	\includegraphics[width=2.4in]{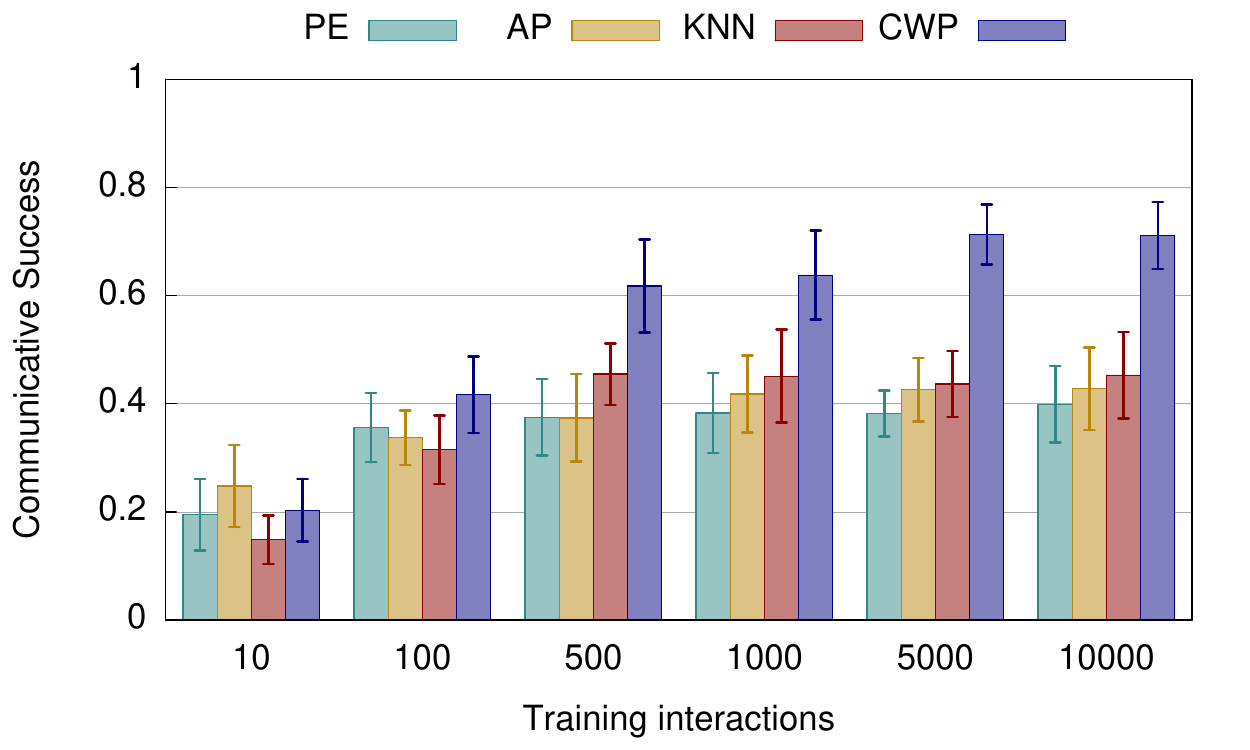}
	\caption{Communicative success for increasing training interactions using the cross-situational learning paradigm ($f=0$). World size = 32, context size = 4, tutor lexicon size = 50, 100 testing interactions, 20 repetitions. Error bars show standard deviation.}
	\label{fig:CSL-dynamics}
\end{figure}

Figure~\ref{fig:CSL-dynamics} compares the communicative success for the four algorithms in a cross-situational learning setting ($f=0$). When comparing the cross-situational learning setting (Fig.~\ref{fig:CSL-dynamics}) to the interactive learning setting (Fig.~\ref{fig:IL-dynamics}), we see that removing the social feedback causes the agents to reach lower communicative success, irrespective of the amount of training. Also, the final communicative success achieved after 10000 interactions lies much lower for the cross-situational learning setting. This shows that the availability of social feedback not only speeds up the learning process, but also improves upon it.

While all algorithms show a decrease in success when removing social feedback, this decrease is much larger for PE and AP when compared to CWP, even though they performed similarly in interactive learning. This is because PE and AP assign equal importance to all objects in the context (and former contexts, in the case of AP) when updating the prototypes. CWP estimates the importance of each object, using the co-occurrence counts, and uses this to update the prototypes. We conclude that CWP can handle the absence of social feedback best.

\subsection{Mixed Feedback Environments ($0 \leq f \leq 1$)}
\label{sec:mixed-learning}

\begin{figure}[!t]
	\centering
	\includegraphics[width=2.4in]{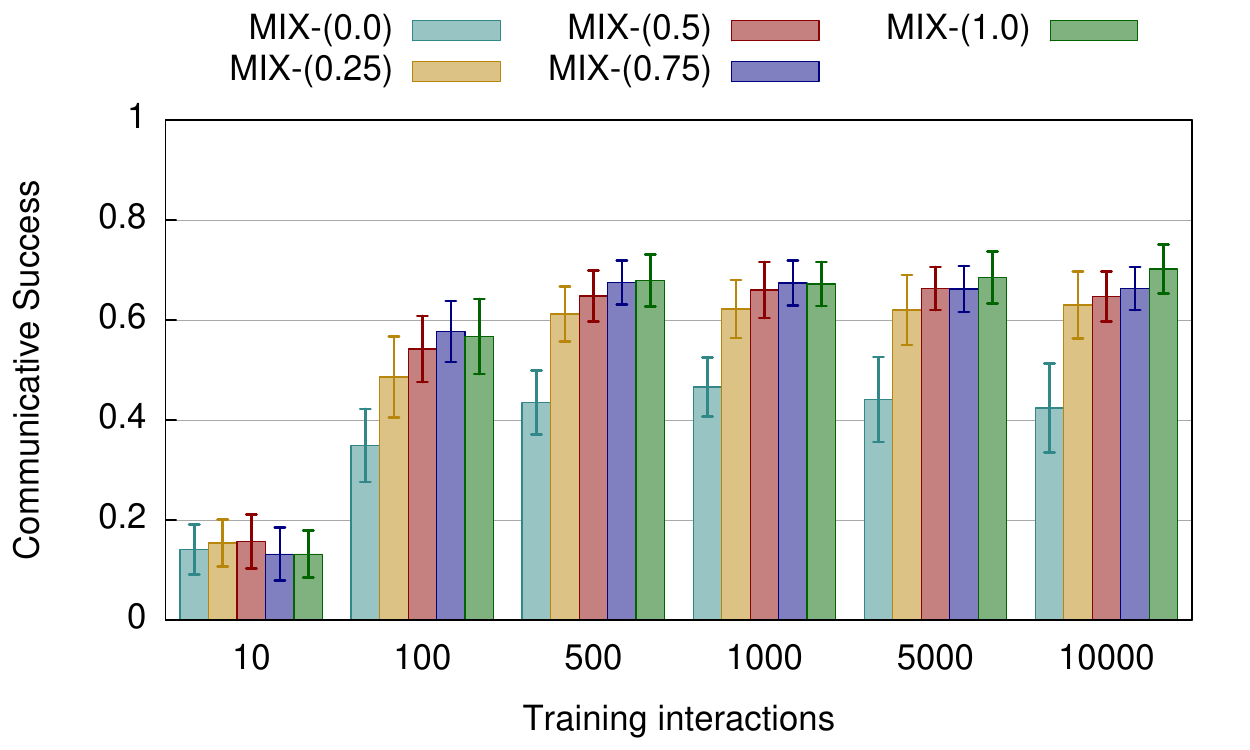}
	\caption{Communicative success for mixed environments using KNN learning algorithm. World size = 32, context size = 4, tutor lexicon size = 50, 100 testing interactions, 20 repetitions. Error bars show standard deviation.}
	\label{fig:MIX-KNN}
\end{figure}

\begin{figure}[!t]
	\centering
	\includegraphics[width=2.4in]{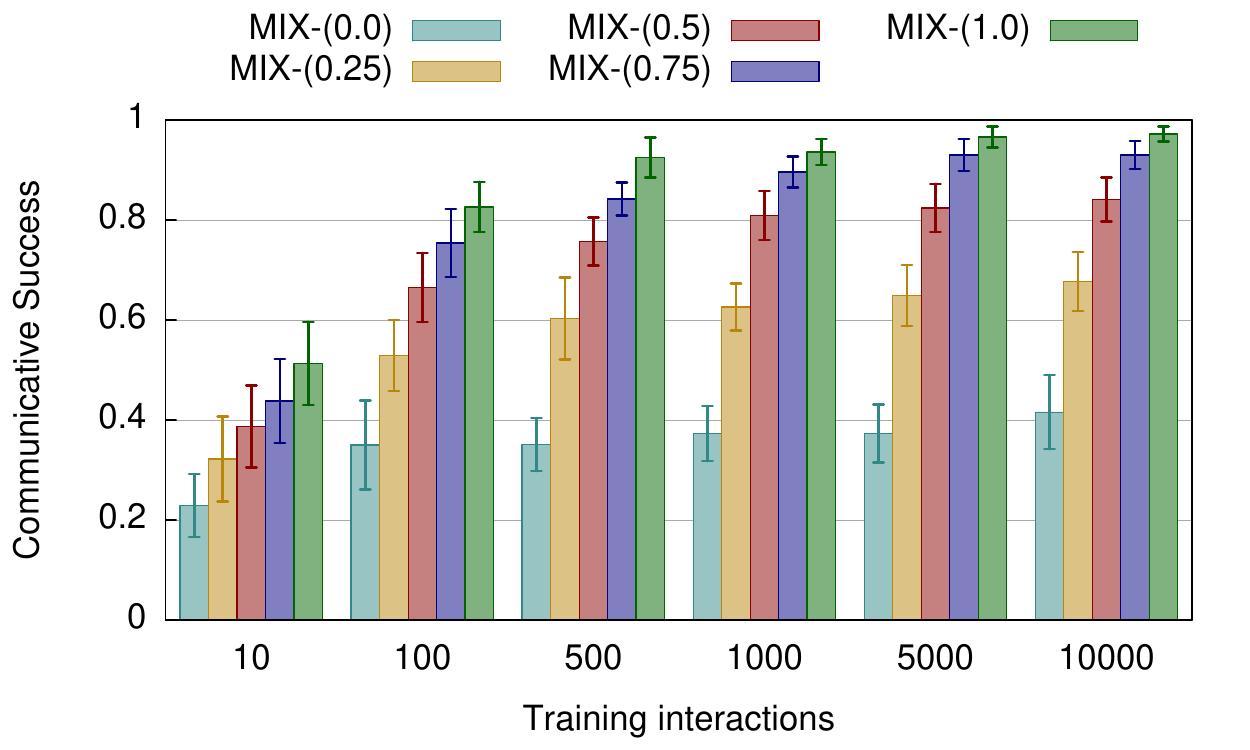}
	\caption{Communicative success for mixed environments using PE learning algorithm. World size = 32, context size = 4, tutor lexicon size = 50, 100 testing interactions, 20 repetitions. Error bars show standard deviation.}
	\label{fig:MIX-PE}
\end{figure}

\begin{figure}[!t]
	\centering
	\includegraphics[width=2.4in]{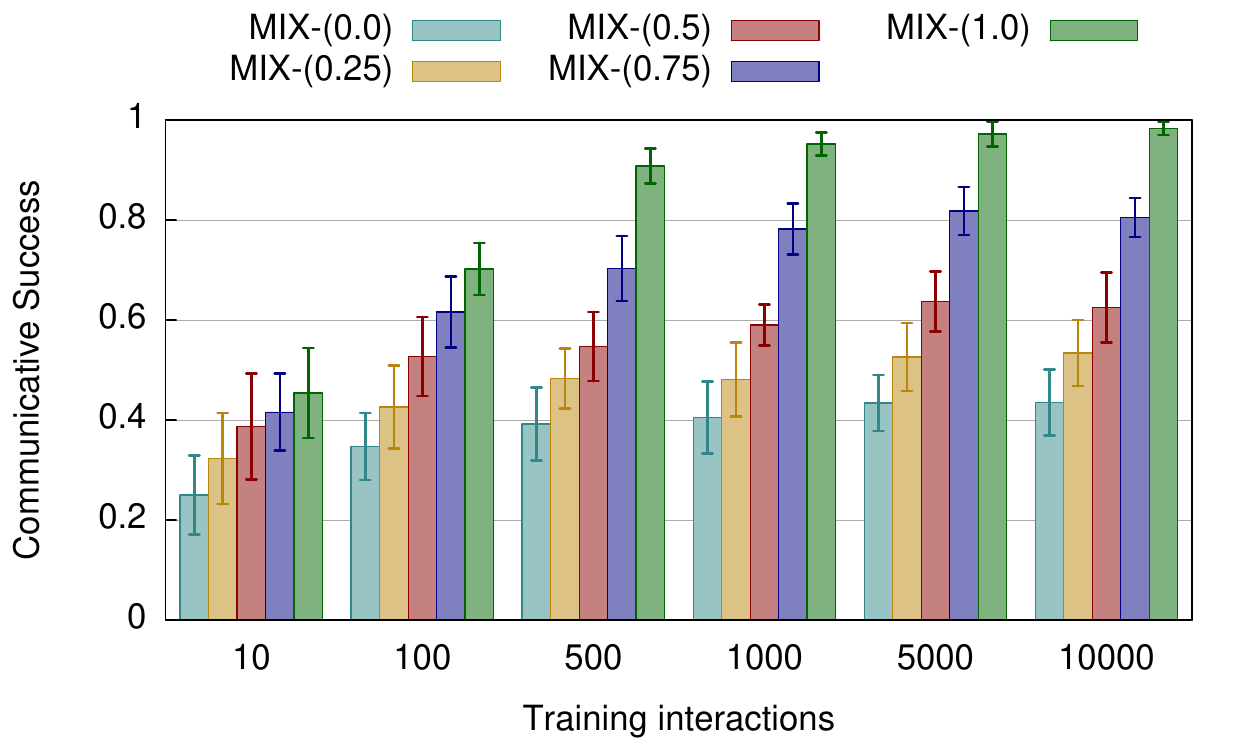}
	\caption{Communicative success for mixed environments using AP learning algorithm. World size = 32, context size = 4, tutor lexicon size = 50, 100 testing interactions, 20 repetitions. Error bars show standard deviation.}
	\label{fig:MIX-AP}
\end{figure}

\begin{figure}[!t]
	\centering
	\includegraphics[width=2.4in]{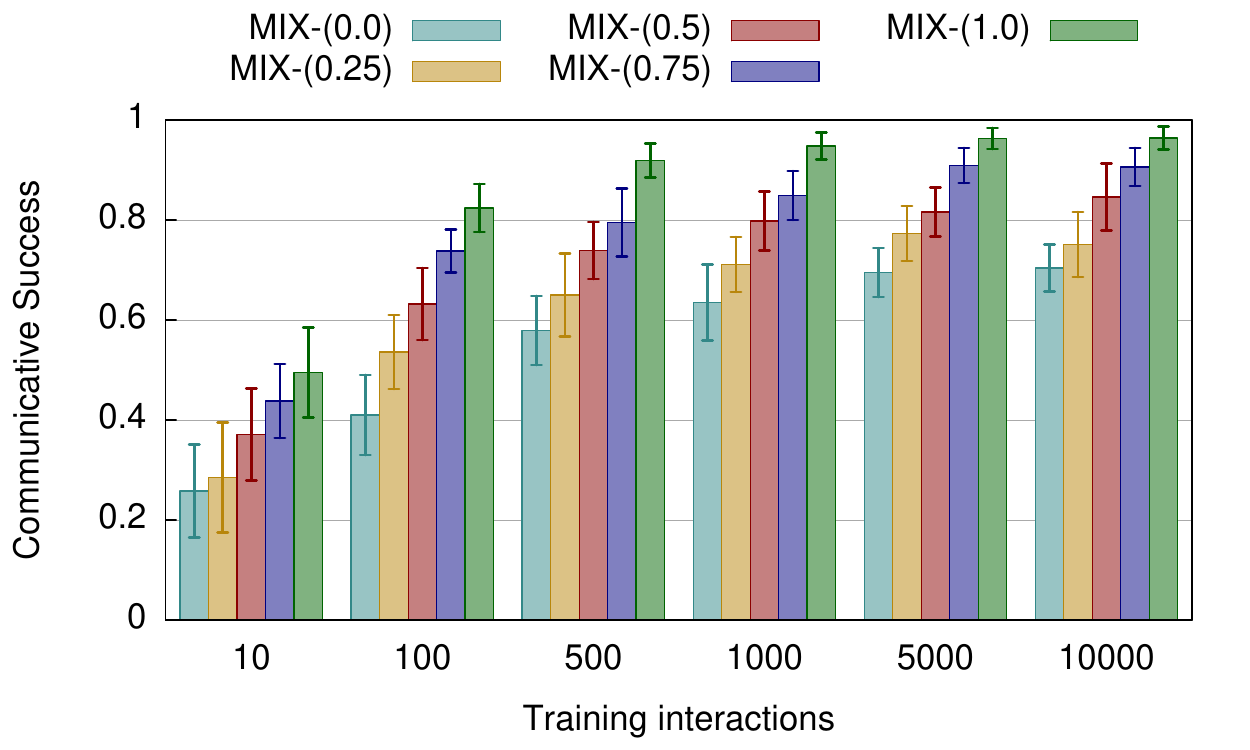}
	\caption{Communicative success for mixed environments using CWP learning algorithm. World size = 32, context size = 4, tutor lexicon size = 50, 100 testing interactions, 20 repetitions. Error bars show standard deviation.}
	\label{fig:MIX-CWP}
\end{figure}

Figures~\ref{fig:MIX-KNN}, \ref{fig:MIX-PE}, \ref{fig:MIX-AP} and~\ref{fig:MIX-CWP} show the communicative success for mixed environments. The behavior of these algorithms for $f=1$ and $f=0$ is already discussed above. For the mixed cases ($0 < f < 1$), the availability of feedback in some interactions does give the learner an advantage. Not only does it reduce the amount of training needed to reach a certain level of success, but also the final performance is higher with increasing social feedback.

While PE and AP perform similarly in the extreme settings ($f=0$ and $f=1$), the PE learner benefits more from the availability of some feedback. For example, the increase in success between $f=0$ and $f=0.25$ is much larger for PE than it is for AP. This is because the presence of feedback allows the PE learner to update its prototype using a single object, while the AP learner still takes previous contexts into account.

Finally, we turn to Fig.~\ref{fig:MIX-KNN}. As discussed above, the KNN algorithm reaches approximately the same communicative success as the other algorithms in the cross-situational learning setting ($f=0$). The availability of some feedback ($0 < f < 1$) immediately increases the success of the learner, close to the performance of interactive learning. However, the performance with full feedback ($f=1$) is not comparable to that of the other algorithms. 

\subsection{Impact Production Strategy}
\label{sec:production-strategy}
One of the parameters in the experimental setup is the tutor strategy for selecting word $w$ given a context $C$ and a topic $o_s$. We assumed what is called a discrimination strategy (Equation \ref{eqn:tutor-prod}) where the tutor selects the most discriminative word - a word that is closest to the topic and most far away from all other objects in the context. Such strategies are commonly used in word learning models (e.g. \cite{spranger2013acquisition})

The tutor can also use another word choice strategy that does not presume the word to be discriminative. This is called \emph{descriptive production strategy} and it is easily described by the following equation where $P$ is the prototypes known to the tutor.
\begin{eqnarray}
\label{eqn:tutor-prod-desc}
p &=& \argmin_{p \in P} \operatorname{d}(p,o_s)
\end{eqnarray}

We test the descriptive production rule in the same settings as discussed before. Everything else, such as learner algorithms, tutor interpretation mechanism (Equation \ref{eqn:tutor-int}), training and test paradigm stays the same.

\begin{figure}[!t]
	\centering
	\includegraphics[width=2.4in]{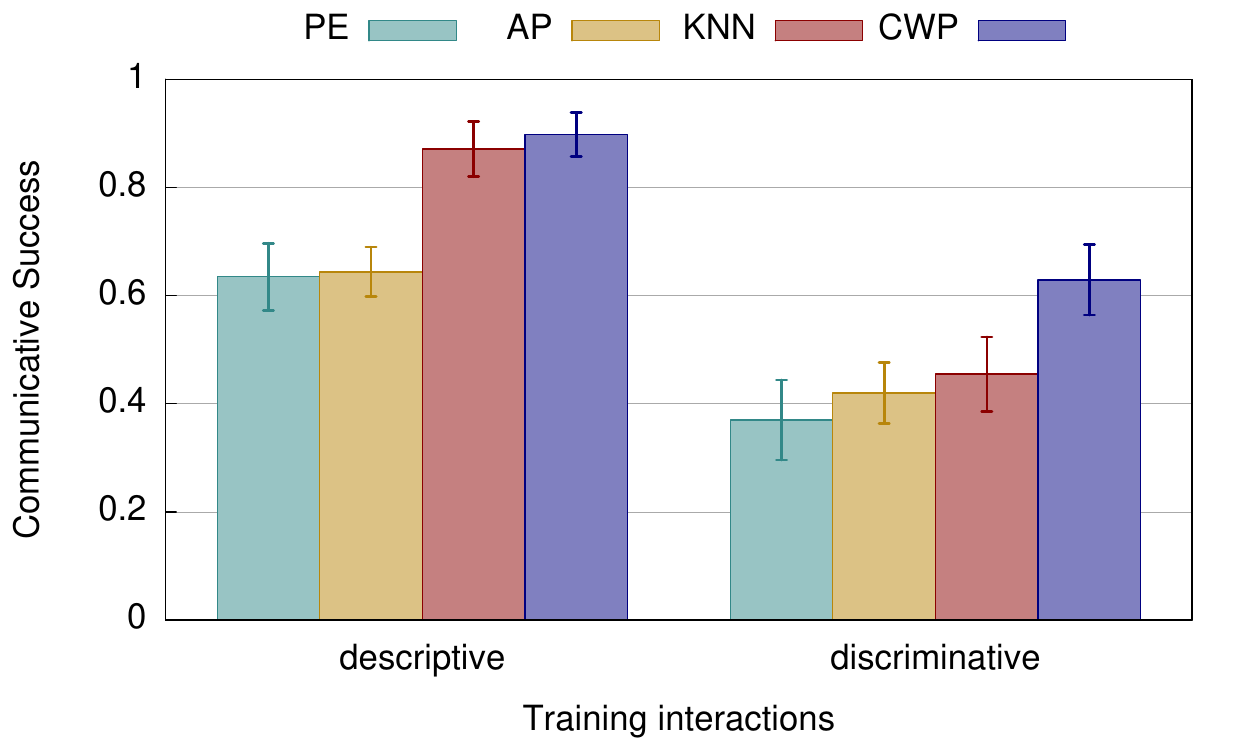}
	\caption{Showing the effect of the tutor's production strategy on the communicative success in cross-situational learning ($f=0$). World size = 32, context size = 4, tutor lexicon size = 50, 1000 training interactions, 100 test interactions, 20 repetitions. Error bars show standard deviation.}
	\label{fig:production-effect-mix0}
\end{figure}

\begin{figure}[!t]
	\centering
	\includegraphics[width=2.4in]{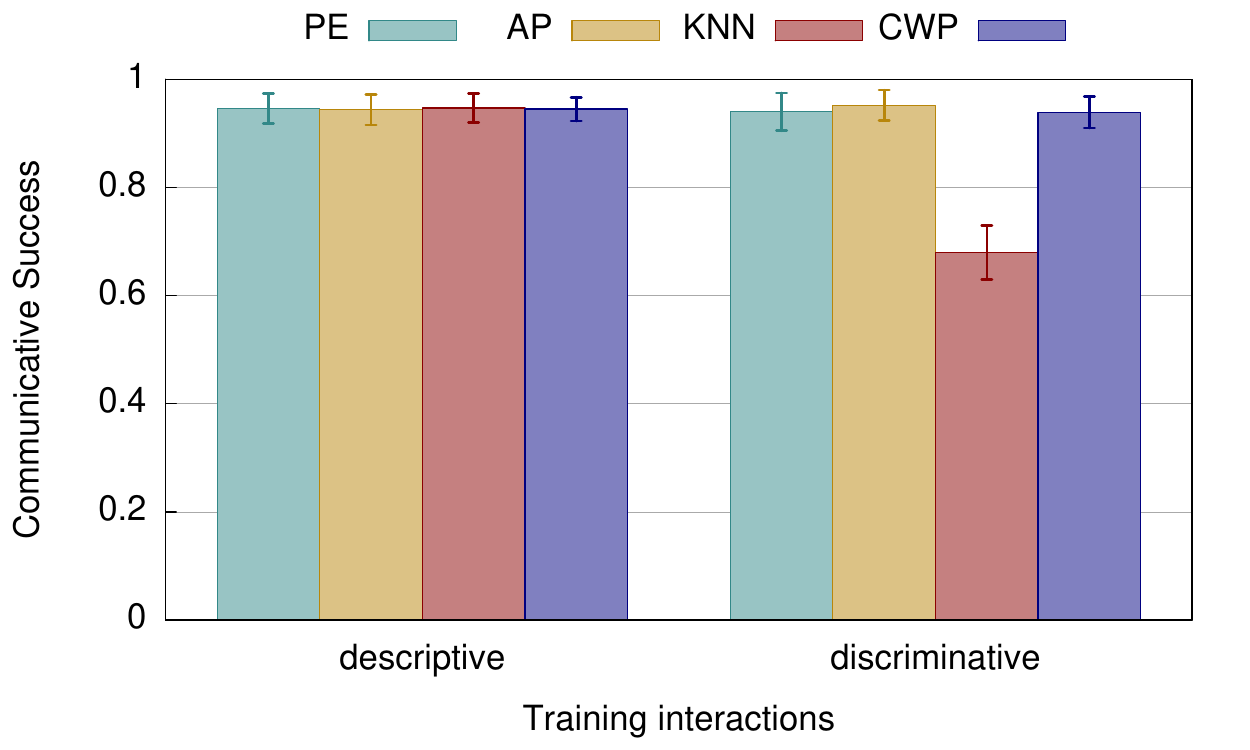}
	\caption{Showing the effect of the tutor's production strategy on the communicative success in interactive learning ($f=1$). World size = 32, context size = 4, tutor lexicon size = 50, 1000 training interactions, 100 test interactions, 20 repetitions. Error bars show standard deviation.}

	\label{fig:production-effect-mix1}
\end{figure}

Figures \ref{fig:production-effect-mix0} and \ref{fig:production-effect-mix1} show the effect of the descriptive production rule on cross-situational learning ($f=0$) and interactive learning environments ($f=1$), respectively.  It is clear that in cross-situational learning environments (Figure \ref{fig:production-effect-mix0}) all algorithms benefit from the descriptive production rule. Similarly, the descriptive production rule also has positive effects on early convergence in interactive learning environments (Figure \ref{fig:production-effect-mix1}).

This is especially true for the KNN algorithm, which is able to reach similar performance as the other algorithms given descriptive tutors. KNN benefits from the descriptive production rule since this rule will always produce the same word for the same object, whereas the discriminative production strategy may more often produce a different word for the same object (because of the influence of distractor objects). KNN exploits the use of the same label in description to better learn the correct labels for each object. The prototype-based algorithms benefit both from singling out the topic from the context (discrimination) and consistent object naming (description).

\section{Conclusion}

Our results show a double effect of tutor feedback on learner success. On the one hand, introducing social feedback in the learning process of the agent allows the agent to acquire the lexicon faster. Across learning algorithms, the learner needs less training to reach the same level of communicative success when there is feedback compared to when there is less or none at all. On the other hand, more feedback also improves the communicative success for the same amount of training. In other words, agents are better at communicating when being trained with more feedback. However, as we have shown in Section~\ref{sec:production-strategy}, this is also dependent on the strategy used by the tutor to provide learning opportunities to the learner. Finally, we can note that our novel algorithms, especially CWP, perform well in the continuous semantic domain -- a domain which has not been explored extensively in the cross-situational learning paradigm.

A next step is to apply these mechanisms to a data set with real robots. This will allow us to further study the effect of tutor feedback in more challenging environments. Indeed, in a grounded environment, each agent has its own view on the objects, causing perceptual deviation. How the agents will respond to this additional difficulty is part of future work.

Similarly is it important to study the quality of the tutor feedback. In the experiments here, the feedback given by the tutor in the form of pointing is handled implicitly. The agents are able to point to the exact centre of an object and they can perfectly interpret the pointing of other agents. In real life, this may not always be the case. Gaze and pointing are noisy reference mechanisms. Therefore, it is interesting to study how well the learner acquires the lexicon in the case of unreliable social feedback.




\bibliographystyle{IEEEtran}
\bibliography{IEEEabrv,references}

\end{document}